\documentclass[final]{cvpr}
\pdfoutput=1
\usepackage{times}
\usepackage{epsfig}
\usepackage{graphicx}
\usepackage{amsmath}
\usepackage{mathtools}
\usepackage{multirow}
\usepackage{adjustbox}

\DeclarePairedDelimiter{\norm}{\lVert}{\rVert} 
\DeclareMathOperator*{\argmin}{arg\,min}

\usepackage{amssymb}

\usepackage{float}
\usepackage{booktabs}
\usepackage{dblfloatfix} 
\usepackage{caption}

\usepackage[dvipsnames]{xcolor}

\usepackage[pagebackref=true,breaklinks=true,colorlinks,bookmarks=false]{hyperref}

\def\cvprPaperID{3333} 
\def\confYear{CVPR 2021}

\begin{document}

\title{Enhanced 3DMM Attribute Control via Synthetic Dataset Creation Pipeline}

\author{

Wonwoong Cho\thanks{These authors contributed equally.}\\
Purdue University\\
\fontsize{11.5pt}{11.5pt}\selectfont
{NAVER WEBTOON}

\and
Inyeop Lee\footnotemark[1] \\
\fontsize{11.5pt}{11.5pt}\selectfont
{NAVER WEBTOON}

\and
David Inouye\\
Purdue University
}

\maketitle


\begin{abstract}
   While facial attribute manipulation of 2D images via Generative Adversarial Networks (GANs) has become common in computer vision and graphics due to its many practical uses, research on 3D attribute manipulation is relatively undeveloped. Existing 3D attribute manipulation methods are limited because the same semantic changes are applied to every 3D face. The key challenge for developing better 3D attribute control methods is the lack of paired training data in which one attribute is changed while other attributes are held fixed---e.g., a pair of 3D faces where one is male and the other is female but all other attributes, such as race and expression, are the same. To overcome this challenge, we design a novel pipeline for generating paired 3D faces by harnessing the power of GANs. On top of this pipeline, we then propose an enhanced non-linear 3D conditional attribute controller that increases the precision and diversity of 3D attribute control compared to existing methods. We demonstrate the validity of our dataset creation pipeline and the superior performance of our conditional attribute controller via quantitative and qualitative evaluations.
\end{abstract}




\section{Introduction}
\label{Introduction}








Facial attribute manipulation on 2D image has drawn significant attention in various computer vision and graphics research, such as GANs~\cite{Shen_2020_CVPR,ganspace,voynov2020unsupervised} and image translation~\cite{CycleGAN2017,choi2018stargan,huang2018munit,GDWCT2019} due to its practical necessity and broad applicability. However, 3D facial attribute manipulation remains relatively unexplored despite its potential impact in many applications including the blendshape techniques~\cite{lewis2010direct,FaceWarehouse}, 2D image manipulation~\cite{tewari2020stylerig,deng2020disentangled,ghosh2020gif}, virtual try-on~\cite{niswar2011virtual} and virtual make-up~\cite{scherbaum2011computer}. 
We posit that this is attributed to a fundamental problem of 3D: the availability of 3D training data is very limited.


Intuitively, the simplest method to achieve the attribute manipulation of 3D faces is to train a regressive model with before-and-after paired data, e.g., a pair of 3D faces for a single identity with different ages (`identity' refers to all other attributes except for an attribute of interest).
However, it is practically difficult to obtain the paired 3D facial data because acquiring 3D scans are expensive and require significant manual labor for  attributes such as facial expression, age, makeup, etc. Moreover, it is fundamentally impossible to collect 3D paired data for certain attributes such as gender, race, facial bone structure, etc.
For example, having all the data of different races with a fixed identity is impossible because the race for each person is unique.

One feasible solution against this seemingly insurmountable obstacle is to make use of Generative Adversarial Networks (GANs). GAN models have been previously verified to enable manipulation of an attribute of an output image while maintaining the identity of the individual by navigating the GANs' latent space.
Inspired by this, we propose to leverage this GAN latent space and analysis-by-synthesis techniques that provide new synthetic 3D facial attribute datasets---which could empower further 3D attribute manipulation research.
In response, we design a pipeline that combines GANs, the GAN manipulation techniques, and 3D reconstruction networks.
This novel pipeline can synthetically generate a large and diverse set of 3D face pairs with attribute annotations based on randomly sampled latent vectors.




The existing methods~\cite{3DMM} in 3D facial attribute manipulation add a single global attribute vector to the vertices of the 3D face input---in particular, the manipulation direction is the same for all possible face inputs.
However, this global additive transformation is limited because it does not consider the characteristics of each individual face.
For example, manipulating the age attribute of a male requires a different additive transformation from that of a female because the conspicuous features of male and female according to different ages are different. e.g., the dominant feature of the young female wears make-up while male wears beard without any make-up.
To this end, on top of the synthetic data creation pipeline we have constructed, we propose a conditional attribute controller which transforms a given 3D face based on the 3D face itself. 


Our contributions can be summarized as follows:
 \begin{itemize}
    \setlength\itemsep{0.5em}
    \item We propose a pipeline to create a novel synthetic dataset for the attribute manipulation of 3D faces.
    \item We will make our synthetic dataset publicly available to accelerate future research on 3D face attribute manipulation.
    \item We develop a novel conditional attribute controller that leverages this new dataset.
    \item We quantitatively and qualitatively demonstrate the potential of the novel synthetic dataset in manipulating 3D face. Moreover, we also verify that our conditional attribute controller clearly improves performance compared to baseline method.
 \end{itemize}

\section{Background and Related Works}
Related works of GANs and 3D Morphable Model (3DMM) are described in subsection~\ref{sub:relatedworks-gans},~\ref{sub:relatedworks-3dmm}. We then provide a technical description on 3DMM in subsection~\ref{subsection:3DMM}. We further explain how the existing method handles the attribute manipulation on top of 3DMM in subsection~\ref{subsection:global-attribute-manipulation}.

\subsection{GANs.}\label{sub:relatedworks-gans}
GANs~\cite{GAN} are arguably widely used generative model. The core idea of GANs is to train a generator in a way that its output distribution matches the data distribution. During the past few years, GANs' performance has been remarkably improved and achieved to produce photo-realistic quality images. ~\cite{WGAN,wgan-gp,miyato2018spectral,brock2018large,Mescheder2018ICML,NEURIPS2018_e46de7e1,Karras_2019_CVPR}.

\vspace{1mm}  \noindent \textbf{attribute manipulation.} Boosted by the enhanced power of GANs, facial attribute manipulation on 2D image has been widely explored. Image translation~\cite{CycleGAN2017,choi2018stargan,huang2018munit,GDWCT2019} is a one research area that aims to translate the facial attribute of a given image to the target attribute. 


Another research direction~\cite{Shen_2020_CVPR,ganspace,voynov2020unsupervised} towards the facial attribute control is to directly manipulate a latent vector on top of the pre-trained latent space. By finding a direction vector for an attribute, those studies proposed to translate latent vectors along the direction, which bring an intended semantic change after forwarding the latent vector into the fixed generator. Advantages of this approach in our frameworks are that the input images are not required and the semantic score described in subsection~\ref{subsection:paired-data-creation} can be easily obtained.

\subsection{3DMM.}\label{sub:relatedworks-3dmm}
Since the introduction of the original 3DMM~\cite{3DMM}, many variants~\cite{1613022,FaceWarehouse,gaussian_mixture_3dmm,LSFM} of the linear statistical model including The Basel Face Model~\cite{bfm09,BFM17} (BFM) have contributed to the improvements of 3DMM. Due to its stable and reliable performance, the 3DMM model has been widely used in diverse fields, such as face recognition~\cite{Face-recognition-based,deepface}, 3D face reconstruction~\cite{zhu2016face,Romdhani05estimating3d,Large-Pose-Face-Alignment,tewari2017mofa} and face reenactment~\cite{kim2018deep,blanz2003reanimating,chai2003vision,face-transfer}. Recently, 3DMM has also been used in 2D image manipulation due to the disentangled nature of the 3D parameters, e.g., light, pose, and expression. Briefly, StyleRig~\cite{tewari2020stylerig} introduced additional networks trained to map the 3D parametric space into the well-trained latent space of GANs. 
DiscoFaceGAN~\cite{deng2020disentangled} proposed an imitative-contrasive learning scheme in which 3D priors are incorporated to achieve the interpretable and controllable latent representations. Meanwhile, GIF~\cite{ghosh2020gif} proposed to exploit the interpretable 3D parameters as a condition for a conditional generative model.

\vspace{1mm}  \noindent \textbf{Intuitive control over the 3DMM parameters.}
For alleviating a problem of scarce semantic interpretation of the 3D parameter, local 3D morphable model~\cite{tena2011interactive,neumann2013sparse} has been explored. Those studies show that a model built from part-based statistics provides the enhanced intuition in manipulating 3D face, making the animation editing via 3DMM more feasible. Another effort towards the intuitive parameter control is from the body models. Briefly, feature-based synthesis~\cite{allen2003space} (e.g., a body model corresponding to the given height and weight is created), and language-based synthesis~\cite{Bodytalk} (e.g., a body model for ``short'' or ``long legs'' is generated.) are presented. Indeed, the technique those body studies are based on is from the original 3DMM paper~\cite{3DMM}. It showed that once a direction vector for a specific attribute in the parametric space is found, manipulating an attribute of a given parameter is possible by shifting the parameter along the direction. Recently, a method~\cite{ghafourzadeh2019part} combining the idea of the local 3D morphable model and the method of the attribute direction vector has been explored.





%
\subsection{Technical background of 3DMM}\label{subsection:3DMM}
A widely-used 3D morphable face model~\cite{3DMM} is a linear statistical model computed from a multivariate normal distribution.  Its shape model is built from the statistics computed over coordinates of each registered 3D scanned data. Given a shape parameter ${p_s}$ that determines the output 3D face, this statistical model can be formulated as $S_{model}(p_s)=\Bar{S}+E_sp_s$, where ${\Bar{S}\in\mathbb{R}^{3n}}$ is a mean shape and ${E_s\in\mathbb{R}^{(3n\times k)}}$ is shape eigenvectors obtained via principal component analysis (PCA) of the 3D scanned data.
${n}$ is the number of vertices and ${k}$ is the number of parameters. Each element in ${p_s\in\mathbb{R}^k}$ is a coefficient of the eigenvectors that determines a single point on the subspace spanned by the eigenvectors. Furthermore, $S_{model}$ can be divided into id and expression model~\cite{FaceWarehouse}, so the shape model we use in our frameworks can be represented as:
\begin{equation}\label{eq:shape-and-expression-model}
    S_{model}(p_i,p_e)=(\Bar{S_i}+\Bar{S_e})+E_ip_i+E_ep_e,
\end{equation} 
where the id-relevant terms ${\{\Bar{S_i},E_i\}}$ are computed from the distribution of 3D scans having neutral expressions, and the expression terms ${\{\Bar{S_e},E_e\}}$ are from the distribution of offsets within expressive and neutral 3D scans.

A texture model is obtained in a similar manner to the shape model, but the statistics are computed along RGB values of each vertex, rather than coordinates, and its formulation is written as:
\begin{equation}\label{eq:texture-model}
    T_{model}(p_t)=\Bar{T}+E_tp_t.
\end{equation} 
Note that the parameters ${\{p_i, p_e, p_t\}}$ are optimized to find a desired 3D face.

\subsection{Global Attribute Manipulation}\label{subsection:global-attribute-manipulation}
Previous studies~\cite{3DMM,Ghafourzadeh2019PartBased3F,Bodytalk,TheSpaceofHumanBodyShapes} have shown that a simple shift in ${p}$ space along a global attribute direction makes a given parameter ${p}$ semantically changed towards the attribute. Briefly, the direction ${\hat{p}}$ is found by solving a rank-1 approximation problem given a matrix of parameters ${P}$ and their corresponding attribute labels ${a}$, i.e., 
\begin{equation}
    \argmin_{\hat{p}} \norm{P-\hat{p}a^T}_F^2,
\end{equation}
where ${\hat{p} \in \mathbb{R}^{k}}$, ${a \in \mathbb{R}^{n}}$, and ${P \in \mathbb{R}^{k\times n}}$.
${k}$ is the number of dimensions of parameter and ${n}$ is the number of data. Once the global direction for the attribute is obtained, an arbitrary parameter ${p}$ is subtracted from or added to ${\hat{p}}$ in order to manipulate the semantics of ${p}$ according to the given attribute, i.e., ${p+s\hat{p}}$, where ${s}$ is a scalar determining the amount of semantic changes.
This leads to a simple linear transformations of the parameters $p$ for altering attributes.

\section{Approach}
In this section, we will concretely describe our entire framework step by step. The overview of the framework is elaborated in subsection~\ref{subsection:overview}. The specific process for obtaining paired data is explained in subsection~\ref{subsection:paired-data-creation}. Lastly, training details will be described in subsection~\ref{subsection:loss-functions}.

\subsection{Overview}\label{subsection:overview}

The ultimate aim of our work is to present a conditional attribute controller that can manipulate any high-level attribute, such as the gender, age, attractiveness and race of a given 3D parameter while keeping other attributes fixed.

Our conditional attribute controller learns to plausibly transform an attribute of a 3D parameter as much as a given score for the attribute. For example, a 3D face of an Asian person can be morphed into any other race, such as white and black, by adjusting the score for the attribute.
Specifically, given an arbitrary parameter ${p}$ with a score ${s}$ for a given attribute, our controller ${f}$ transforms the given ${p}$ to the ${\hat{p}}$, i.e.,
\begin{equation}
    \hat{p}=f(p, s),
\end{equation}
where ${f}$ is designed to be neural networks in our work. The output ${\hat{p}}$ is then added to the original ${p}$, so that a transformed parameter ${\tilde{p}}$ is formulated as
\begin{equation}
    \tilde{p}=p+\hat{p}.
\end{equation}

\begin{figure}[t]
\vspace*{-0.5cm}
  \includegraphics[width=\linewidth]{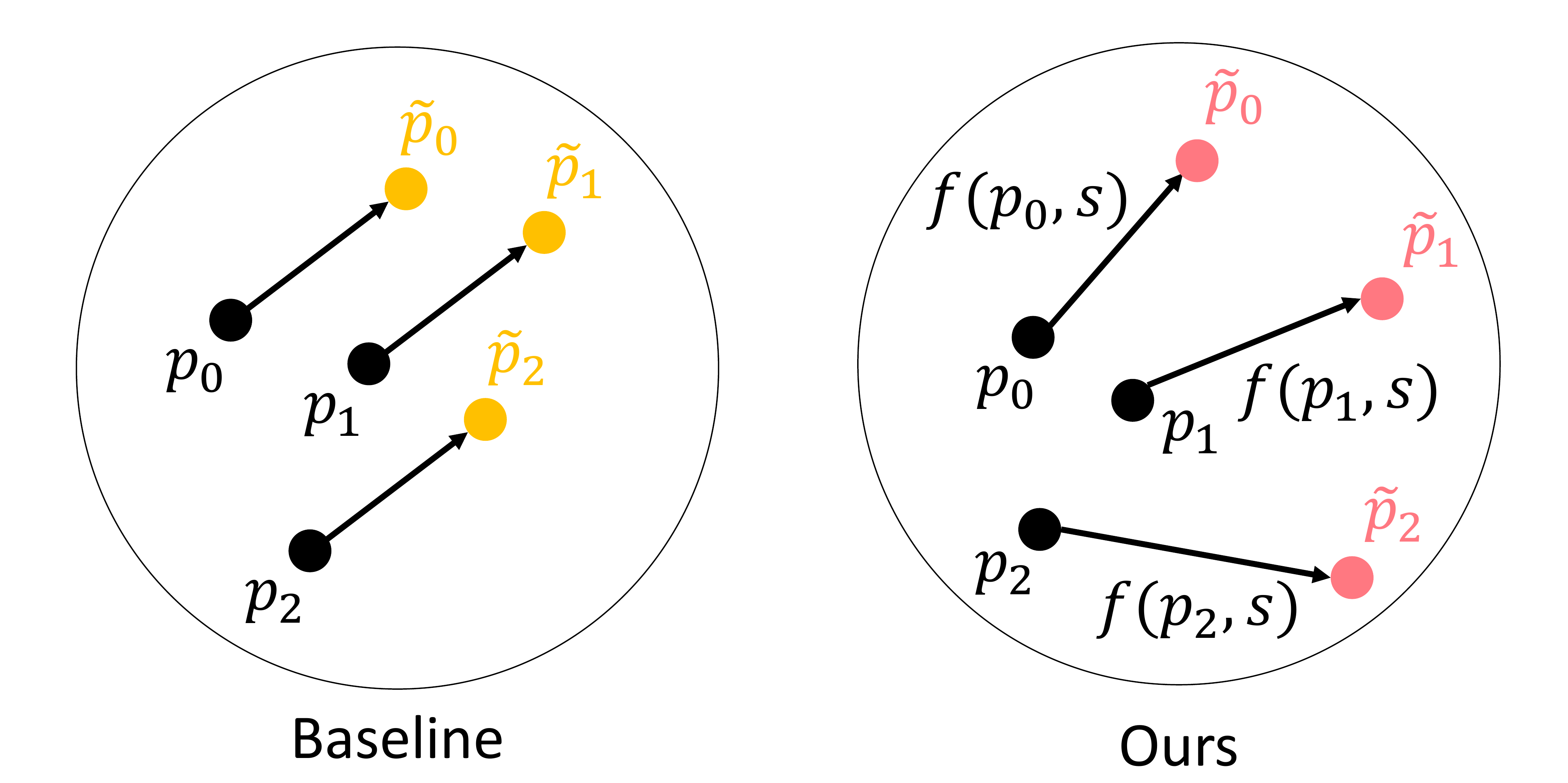}
  \caption{While the baseline model only produces additive shifts for each feature, our proposed approach can modify the face parameters non-linear ways for different faces, e.g., the semantic transformation of a White into an Asian is different from that of a Black into an Asian.}
\label{fig:conceptual-figure}
\end{figure}

Note that the intended result of ${\hat{p}}$ is a conditional attribute direction on the parametric space with a proper norm, and ${\tilde{p}}$ is the transformed parameter that properly reflects the desired attribute transfiguration.

Intuitively, as seen in a conceptual illustration in Fig.~\ref{fig:conceptual-figure}, the transformed parameter ${\tilde{p}}$ through our conditional attribute controller is $\tilde{p} = p + f(p,s)$, where $f$ could be non-linear with respect to $p$. However, ${\tilde{p}}$ through the baseline is $\tilde{p} = p + s\hat{p}$, which is a linear transformation in ${p}$ space. This key difference brings a superior performance of our model over the baseline.
This could also be seen as a residual network structure where the original input is added to the output of the network. 
We followed this scheme because predicting the residual is simpler than directly producing the new parameter, which would mean simply $\tilde{p} = f(p,s)$.
In our experiments, we empirically verify that the residual structure generally estimates better transformed parameters.

\begin{figure*}[t]
  \includegraphics[width=\linewidth]{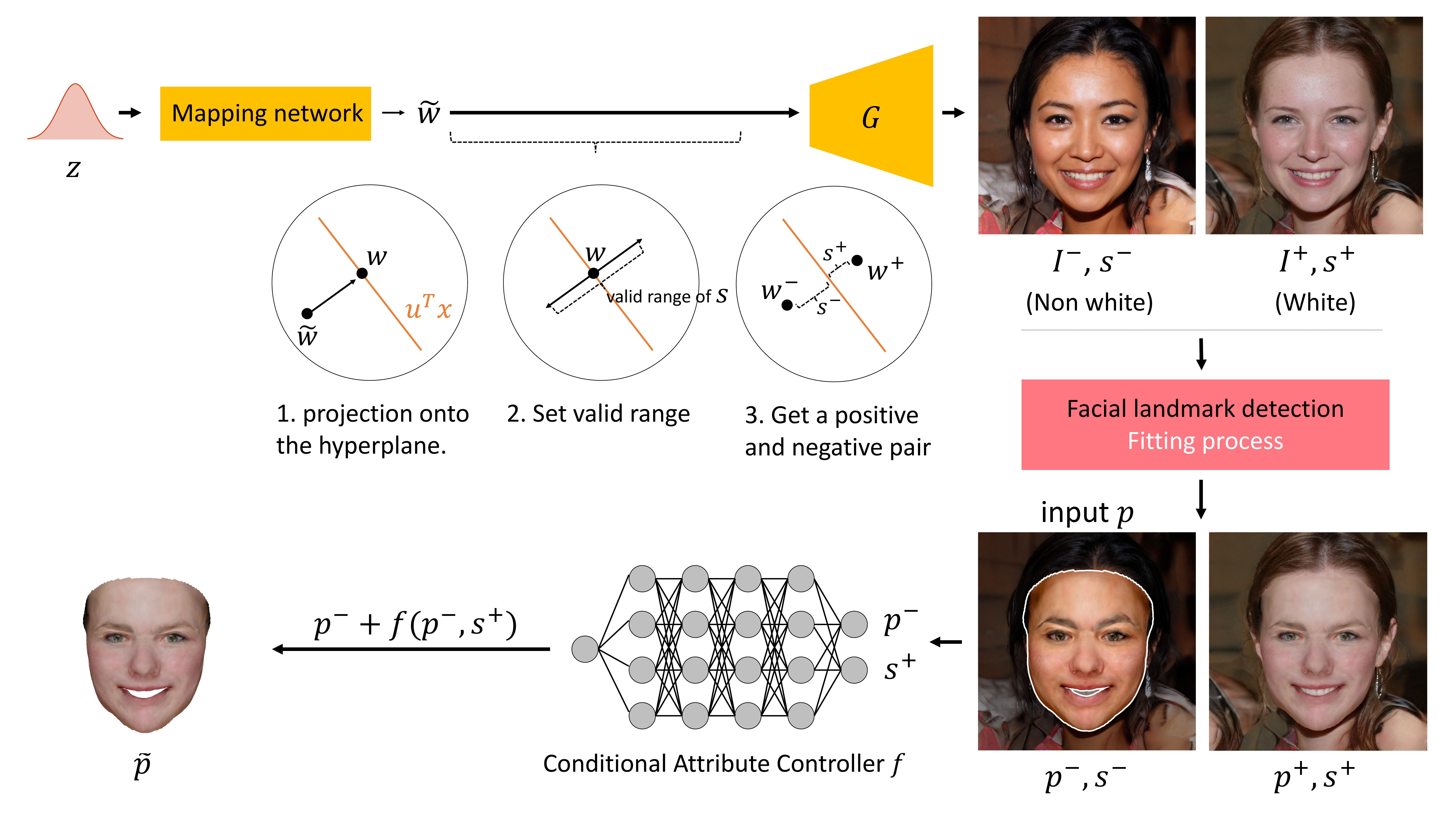}
  \caption{The overview of our novel pipeline. On top of the well-defined latent space of StyleGAN~\cite{Karras_2019_CVPR}, we first sample paired data and its semantic scores by utilizing the hyperplane in the latent space. Once facial landmarks for the sampled images are gained, we find the best 3DMM parameters by a method in analysis-by-synthesis. By leveraging the parameters and the semantic scores as training data, we train our conditional attribute controller ${f}$ which learns to estimate the best attribute transformation for a given parameter.}
\label{figure:main-fig}
\end{figure*}

\subsection{Synthetic Dataset Creation}\label{subsection:paired-data-creation}
In order to provide a supervision for the conditional attribute controller, the paired data is required. In this subsection, we present our novel pipeline for creating the synthetic dataset. The pipeline is based on studies on the semantic manipulation on top of GAN latent space and 3D reconstruction.

\subsubsection{Semantic Navigation on GAN Space}
Throughout numerous studies~\cite{Shen_2020_CVPR, icml2020_2025, tewari2020stylerig, shen2020closedform}, GAN latent space has been demonstrated as a well-defined space in which facial attributes are disentangled; a simple shift on the latent space along a specific attribute direction brings about a modification of the attribute while an identity is maintained. Inspired by this, we build up a start point of our pipeline. In order for acquiring a high-quality paired images, we adopt StyleGAN~\cite{Karras_2019_CVPR} to our frameworks. 

\vspace{1mm}  \noindent \textbf{Paired Data Acquisition.} As Fig.~\ref{figure:main-fig}  illustrates, on top of the pretrained networks, we first find a hyperplane on the GAN latent space separating a binary class, e.g., male vs. female or old vs. young, etc., meaning we have its normal vector as well. Our aim in this part is to have a positive and a negative pair of images for an attribute. 

We first project a randomly sampled latent vector ${\tilde{w}\in\mathbb{R}^{d}}$ onto the hyperplane, where $\tilde{w}$ is a latent vector after a mapping network in StyleGAN~\cite{Karras_2019_CVPR}, i.e., $w = {\tilde{w}-proj_u(\tilde{w})} = \tilde{w}-(\tilde{w}^Tu)u$,
where ${u}$ is a unit normal vector of the hyperplane. Then using the normal vector as our direction vector of the attribute, we shift ${w}$ along the direction as much as given score ${s}$, i.e., ${\hat{w}=w+su}$, where ${s}$ determines the extent of a semantic transfiguration, and each of ${\hat{w}\in\mathbb{R}^{d}}$ is a semantically transformed latent vector.

Specifically, we sample a positive and a negative sample with \emph{the same identity} to make a paired data for each given projected latent ${w}$.
It is critical that each element of the pair has the same identity for our training (i.e., all other attributes are held fixed except the target attribute). In order for this, we first set a maximum range of ${s}$ for each attribute, e.g., from -2 to +2 for the white attribute and from -3 to +3 for the black attribute.
\footnote{As done in InterFaceGAN~\cite{Shen_2020_CVPR}, we empirically set the sampling range. Based on an observation, the valid range for each attribute is different, and outside the range, the number of failure cases is seriously increased. 
We expect this is caused because only subspace of  StyleGAN~\cite{Karras_2019_CVPR} are trained well and thus, $w$ outside of this subspace will not produce the expected images.

We then uniformly sample two ${s}$ values, one for a positive and the other for a negative sample.} This process enables us to have the infinite number of paired data with a low price, at which point this has a great potential in 3D research area, where acquiring dataset is expensive. Lastly, by forwarding those manipulated ${w}$s into the generator, we acquire the paired images, which can be represented as ${\{((I,s)^{+}, (I,s)^{-}))_0,...,((I,s)^{+}, (I,s)^{-}))_n\}}$. Note that ${w}$ can be included in our dataset as well, if necessary~\cite{tewari2020stylerig}.


\vspace{1mm}  \noindent \textbf{Semantic score.}
With regard to the necessity of the score in our synthetic dataset, it is unclear that how much each ${w}$ has a specific feature for the attribute. For example, given a set of ${w}$s on the same side from the `young' boundary, it is possible that some ${w}$s are included in baby and the other are in teenager, of which marked features are different. We posit that this huge intra-variation within a same class may puzzle the networks to properly learn. Based on this insight, the semantic score, which is a vector norm between ${w}$ and the hyperplane is included in our dataset.


\subsubsection{3DMM fitting}
Our focus lies in the 3D face manipulation, thus obtaining a shape, expression, and texture parameters ${p=[p_i,p_e,p_t]}$ for the pairs of data is another important step in our pipeline. For the sake of this, reconstruction of 3D parameters from a single image is required. Following the analysis-by-synthesis technique~\cite{History}, we perform our fitting process. Specifically, we acquire the facial landmarks~\cite{Kazemi_2014_CVPR} for each image. We then adopt one of the off-the-shelf 3D reconstruction networks~\cite{deng2019accurate} to our pipeline to obtain an initial 3DMM parameter for each image. Lastly, we follow an optimization-based 3DMM fitting method~\cite{Blanz_fitting,Estimating3DShape,History} for monocular face reconstruction, with which the initial parameter is iteratively optimized. 

\vspace{1mm}  \noindent \textbf{Setup.}
Our 3D face model is composed of 2009 Basel Face Model~\cite{bfm09} for the id and the texture models in Eq.~\ref{eq:shape-and-expression-model},~\ref{eq:texture-model} and the expression model in Eq.~\ref{eq:shape-and-expression-model} built from FaceWarehouse~\cite{FaceWarehouse}. Regarding the image formation process, the perspective camera with an empirically-set focal length is used in order to project a 3D face onto the image plane, and the illumination model is built upon phong shading~\cite{phong}. Each parameter has a dimension of ${\{p_s,p_t\}\in\mathbb{R}^{80}}$,${p_e\in\mathbb{R}^{64}}$. The camera parameters are composed of extrinsic camera parameters, i.e., ${p_c=[x_{R},y_{R},z_{R},x_{T},y_{T},z_{T}]}$, and the light parameters are composed of ${p_l=[x_l,y_l,z_l,r_a,g_a,b_a]}$, where the subscript ${l}$ refer to the light location and the subscript ${a}$ indicate ambient colors.

\vspace{1mm}  \noindent \textbf{Losses for 3DMM fitting.}
For fitting a 3DMM parameter for an image, we adopt the previous techniques in analysis-by-synthesis.
In particular, fitting the 3DMM parameters to a given image can be accomplished by minimizing a combination of energy functions that measure pixel-wise and feature-based error, respectively. The pixel-level loss can be written as:
\begin{equation*}
    E_{pixel}=\sum_{(x,y)\in \mathcal{F}}\norm{I_{trg}(x,y)-I_{render}(x,y)},
\end{equation*}
where ${\mathcal{F}}$ is a foreground region (i.e., the face region) and ${I_{render}}$ is a rendered image, which can be represented as $\mathcal{R}(\Pi(S_{model}(p_i,p_e),p_c),T_{model}(p_t),p_l)$), where ${\Pi}$ is a camera projection model, and ${\mathcal{R}}$ indicates a rendering function~\cite{liu2019softras} including lighting and rasterization process. 
On the other hand, the feature-based energy is measured by a comparison between facial landmarks of the target image and the corresponding 2D-projected vertices, i.e., 
\begin{equation*}
E_{feature}=\mathbb{E}\norm{t_{trg}-t_{proj}},
\end{equation*}

where each of ${\{t_{trg},t_{proj}\}\in\mathbb{R}^{68\times 2}}$ is the 2D landmark coordinates of the target image and a subset of the projected vertices.\footnote{The vertex indices to be included in the subset is predefiend.} Concisely, ${t_{proj}}$ can be obtained via the camera projection model, i.e., $t_{proj}={\Pi(S_{model}(p_i,p_e),p_c)}$.

As a result, a 3DMM parameter ${p}$ for each image is also included in our synthetic dataset, which can be represented as ${\{((s,p)^{+}, (s,p)^{-})_0}$, ..., ${((s,p)^{+}, (s,p)^{-})_n\}}$.

\subsection{Loss Functions for Attribute Controller}\label{subsection:loss-functions}

Once the synthetic dataset is created, we can make use of a paired data for training our conditional attribute controller. For the brevity, we describe the loss functions with a single paired data, i.e, ${((s^+,p^+), (s^-,p^-))_0}$. 

Let a source parameter and a score be ${p_{src},s_{src}}$ and targets be ${p_{trg},s_{trg}}$. Each of positive and negative pairs is randomly set to be the source or the target. The aim of a training process is to make our controller properly learn the semantic transformation. For accomplishing the aim, two objectives are required to be fulfilled.
First, given a source parameter $p_{src}$ and a target score ${s_{trg}}$, our controller ${f}$ has to output ${\hat{p}}$ that makes ${\tilde{p}}$ whose identity is similar to $p_{src}$ but the semantic score is equivalent to ${s_{trg}}$.
This objective can be formulated as:
\begin{equation}
    L_{trg}=\norm{p_{trg}-(p_{src}+f_\theta(p_{src},s_{trg}))}.
\end{equation}
Note that by construction of our paired dataset, the identity of $p_{src}$ and $p_{trg}$  are the same (except for the manipulated attribute) and thus $f_\theta$ learns how to transform the target attribute only.


\begin{table*}[ht]
\begin{center}
\begin{adjustbox}{width=\columnwidth*2,center}
\begin{tabular}{c|ccccccccccccccccc}

\toprule

 & Asian & Attractive & B.L. & Black & B.E. & Chubby & H.C. & Hispanic & Indian & Makeup & Male &  N.E. & NoBeard & P.N. & R.C. & White & Young\\
\midrule
\begin{tabular}{@{}c@{}}Baseline\end{tabular} 
& 7.03 & 14.04 & 9.67 & 10.44 & 9.13 & 9.12 & 10.28 & 8.98 & 8.36 & 7.51 & 14.11 & 10.57 & 8.83 & 9.29 & 9.66 & 5.89 & 12.43\\
\midrule
\begin{tabular}{@{}c@{}}Ours \\w.o.\textbf{res}\end{tabular} 
& \textbf{6.91} & \textbf{11.65} & 9.16 & 9.7 & 8.81 & 8.9 & 9.47 & 8.98 & 8.37 & 7.97 & 11.9 & \textbf{9.61} & 9.77 & 9.12 & 9.39 & 6.65 & \textbf{11.14}\\
\midrule
\begin{tabular}{@{}c@{}}Ours\end{tabular} 
& 7.51 & 11.93 & \textbf{8.98} & \textbf{9.61} & \textbf{8.57} & \textbf{8.64} & \textbf{9.39} & \textbf{8.75} & \textbf{8.06} & \textbf{7.42} & \textbf{12.17} & 9.56 & \textbf{9.36} & \textbf{8.86} & \textbf{9.27} & \textbf{5.93} & 11.24\\
\bottomrule
\end{tabular}
\end{adjustbox}
\end{center}
\vspace*{-0.5cm}
\caption{L2 distance for various attributes averaged over 5 cross validation folds. ``w.o.res'' denotes our method without a residual structure (see subsection~\ref{subsection:overview}). B.L., B.E., H.C., N.E., P.N., and R.C. indicate Big Lips, Bushy Eyebrows, High Cheekbones, Narrow Eyes, Pointy Nose, and Rosy Cheeks. }
\label{table:l2}
\vspace*{-0.3cm}
\end{table*}
\section{Experiments}
In this section, we detail the experimental settings and results. First, implementation details omitted in the previous section are provided in subsection~\ref{subsection:implementation-details}. Second, quantitative and qualitative experiments are elaborated in subsections~\ref{subsection:quantitative evaluations}~and~\ref{subsection:qualitative evaluations}.

\subsection{Implementation Details for the Pipeline}\label{subsection:implementation-details}
\vspace{1mm}  \noindent \textbf{Training details for GANs.}
We used the pre-trained weights of StyleGAN downloaded from an official GitHub repository, which is trained with the Flickr-Faces-HQ (FFHQ) dataset~\cite{Karras_2019_CVPR}.
The dataset consists of 70,000 high-quality images of 1024 resolution.
We adopted the combination of StyleGAN and FFHQ for our pipeline (out of numerous GAN architectures and several high-quality datasets~\cite{karras2018progressive}) because this combination has a semantically disentangled GAN latent space.
In particular, as demonstrated in the paper~\cite{Karras_2019_CVPR}, the latent space of StyleGANs trained with FFHQ is well-defined and linearly separable. Thus, we can easily generate paired data.

\vspace{1mm}  \noindent \textbf{Estimating attribute hyperplanes in GAN latent space.}
Given labeled samples in the GAN latent space, we can estimate an attribute hyperplane separating two classes by fitting a linear support vector machine (SVM)~\cite{Karras_2019_CVPR,Shen_2020_CVPR}.
To create labeled samples (in the latent space), we first generate the image corresponding to the latent sample and then classify the image using a pre-trained image classifier---thus, we can obtain a label for any latent vector.
The image classifier is trained with the CelebA dataset~\cite{liu2015faceattributes}, which has labels for 40 facial attributes, e.g., `Young',`Male', and `Attractive', and the UTKFace dataset~\cite{zhifei2017cvpr}, which has labels for race, e.g., `White',`Black',`Asian', etc.
After acquiring the hyperplane for an attribute, we sample the paired data following the process concretely described in subsection~\ref{subsection:paired-data-creation}.




\subsection{Quantitative Evaluations}\label{subsection:quantitative evaluations}


\vspace{1mm}  \noindent \textbf{Comparisons on the synthetic dataset.}
In order to demonstrate the benefits of the paired data, we compare the L2 distance in the 3DMM parameter space on our synthetic dataset (which is taken as the ground truth in this experiment).
For rigorous comparison, we use 5-fold cross validation, i.e., each fold is composed of 4000 test samples and 16000 train samples.
The final L2 distance is the average over 5 folds.

As seen in Table~\ref{table:l2}, our conditional attribute controller outperforms the baseline model for every attribute indicating that leveraging paired data improves performance in attribute manipulation.
In Table~\ref{table:l2}, we also show the effectiveness of the residual structure in our controller.
We observe performance gain in most of the attributes. Note that our method still outperforms the baseline in a majority of cases even without the residual structure.
This indicates that our conditional attribute controller trained with paired data estimates better semantic transformations for a given 3DMM parameter and attribute score than the baseline model, which globally transforms a parameter regardless of the input 3DMM parameter and score.

\vspace{1mm}  \noindent \textbf{Our controller trained with our synthetic dataset \emph{v.s.} the baseline learned from real dataset.}
We further conduct an experiment for verifying the combination of our synthetic dataset and our conditional attribute controller in practice.
However, two practical issues exist to directly compare the combination to the real dataset.
First, large-scale 3D datasets that have diverse attribute labels do not exist to the best of our knowledge.
Second, real 3D data does not contain paired samples (e.g., a transformed parameter from female to male does not have ground-truth), which makes evaluation difficult. 

In response to the first issue, we make use of FFHQ~\cite{Karras_2019_CVPR} dataset, of which samples for each attribute is sufficiently large.
We obtain 3D parameters using the analysis-by-synthesis technique and the attribute labels of the images through pre-trained classifiers.
For the evaluation metric, we measure the average Mahalanobis distance between the transformed parameters from our controller and the distribution of the FFHQ parameters, i.e., $\frac{1}{N}\sum_{p\in N} \sqrt{(p-\mu)^{T} S^{-1} (p-\mu)}$, where $N$ is the number of the transformed parameters, and ${\mu}$ and ${S}$ are the mean and covariance matrix of the FFHQ parameters, respectively.
The Mahalanobis distance is equivalent to the number of standard deviations between the parameter ${p}$ and the mean ${\mu}$.
Intuitively, the closer a transformed parameter is to the FFHQ distribution, the smaller the distance between them will be.

\begin{table*}[t]
\begin{center}
\begin{adjustbox}{width=\columnwidth*2,center}
\begin{tabular}{c|ccccccccccccccccc}

\toprule

 & Asian & Attractive & B.L. & Black & B.E. & Chubby & H.C. & Hispanic & Indian & Makeup & Male &  N.E. & NoBeard & P.N. & R.C. & White & Young\\
\midrule
\begin{tabular}{@{}c@{}}Baseline \\ (FFHQ)\end{tabular} 
& 14.62 & 14.26 & 14.31 & 14.37 & 14.09 & 14.24 & 14.49 & \textbf{4.87} & 19.78 & 14.64 & 14.58 & 14.04 & 14.7 & 15.93 & 18.43 & 13.94 & 14.3\\
\midrule
\begin{tabular}{@{}c@{}}Ours \\ (synthetic)\end{tabular} 
& \textbf{13.43} & \textbf{12.01} & \textbf{12.63} & \textbf{12.64} & \textbf{12.42} & \textbf{12.52} & \textbf{12.47} & 5.8 & \textbf{17.84} & \textbf{13.47} & \textbf{12.42} & \textbf{11.98} & \textbf{13.14} & \textbf{14.29} & \textbf{16.71} & \textbf{12.94} & \textbf{12.16}\\
\bottomrule
\end{tabular}
\end{adjustbox}
\end{center}
\vspace*{-0.5cm}
\caption{The Mahalanobis distance between the output parameters and the FFHQ distribution.
The abbreviations are the same with Table~\ref{table:l2}.
Each distance is the average over forward and backward attribute transformations, e.g., we average over two distances, one for the distance between the transformed male and the FFHQ male distribution and the other for a distance between the transformed female and FFHQ female distribution. 20\% of FFHQ parameters are used as the test set and the other 80\% are used to train the baseline method.}
\label{table:mahalanobis}
\vspace*{-0.3cm}
\end{table*}

As shown in Table~\ref{table:mahalanobis}, our conditional attribute controller shows the superior performance over the baseline method in terms of the Mahalanobis distance. Note that our conditional attribute controller learns from only the synthetic dataset of parameters while the baseline method learns from the real FFHQ parameters.
This experiment demonstrates both the promising power of our novel synthetic dataset and the superior performance of our conditional attribute controller.

\subsection{Qualitative Evaluations}\label{subsection:qualitative evaluations}

\begin{figure}
  \includegraphics[width=\linewidth]{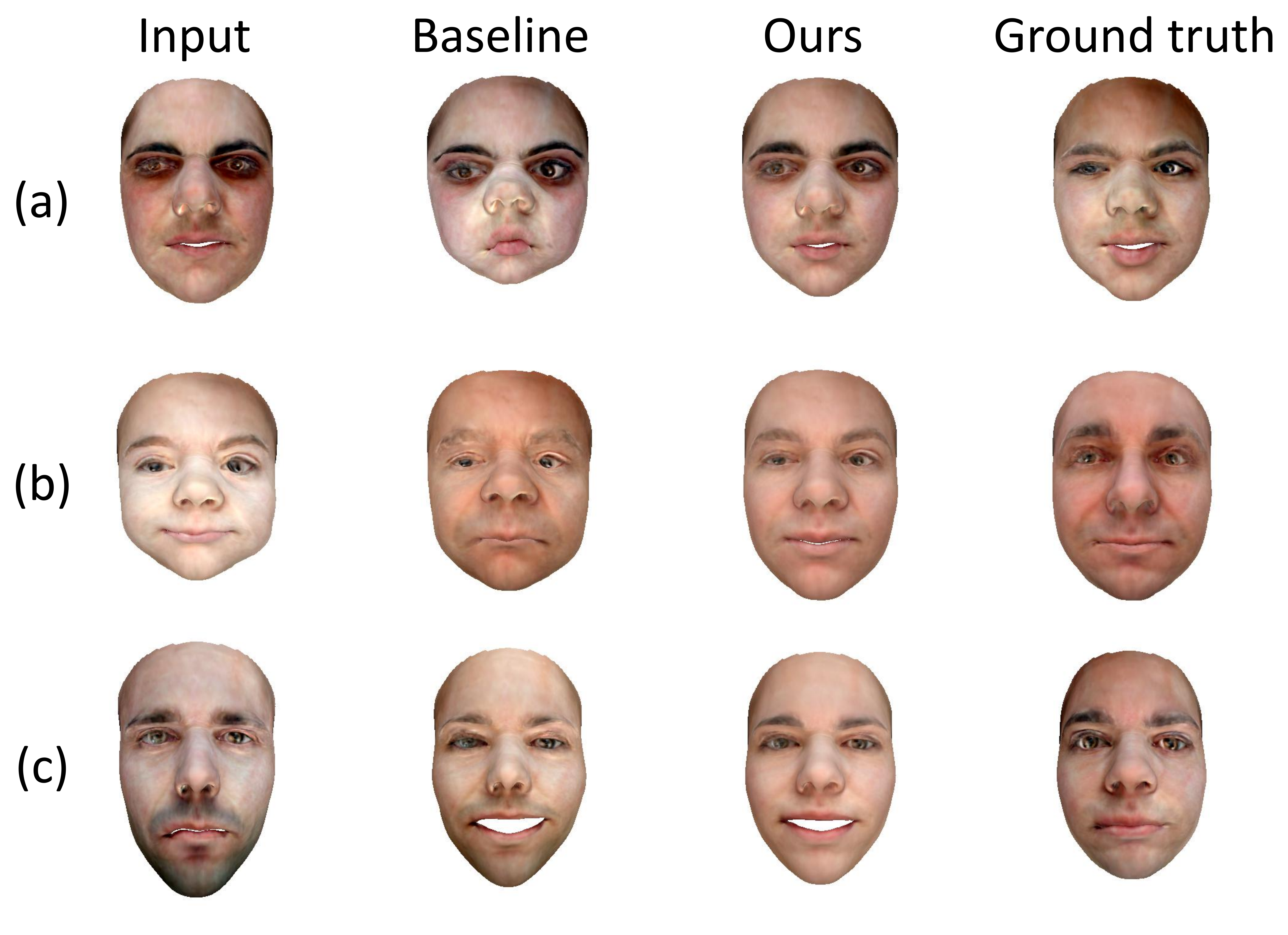}
  \caption{Visualization of samples used in Table~\ref{table:l2}. (a) denotes `Old'${\rightarrow{\text{`Young'}}}$, (b) is `Attractive'${\rightarrow{\text{`NotAttractive'}}}$ and (c) indicates `Male'${\rightarrow{\text{`Female'}}}$.} 
\label{fig:l2-comparisons}
\vspace*{-0.3cm}
\end{figure}

\begin{figure*}[t]
\vspace*{-0.5cm}
  \includegraphics[width=\linewidth]{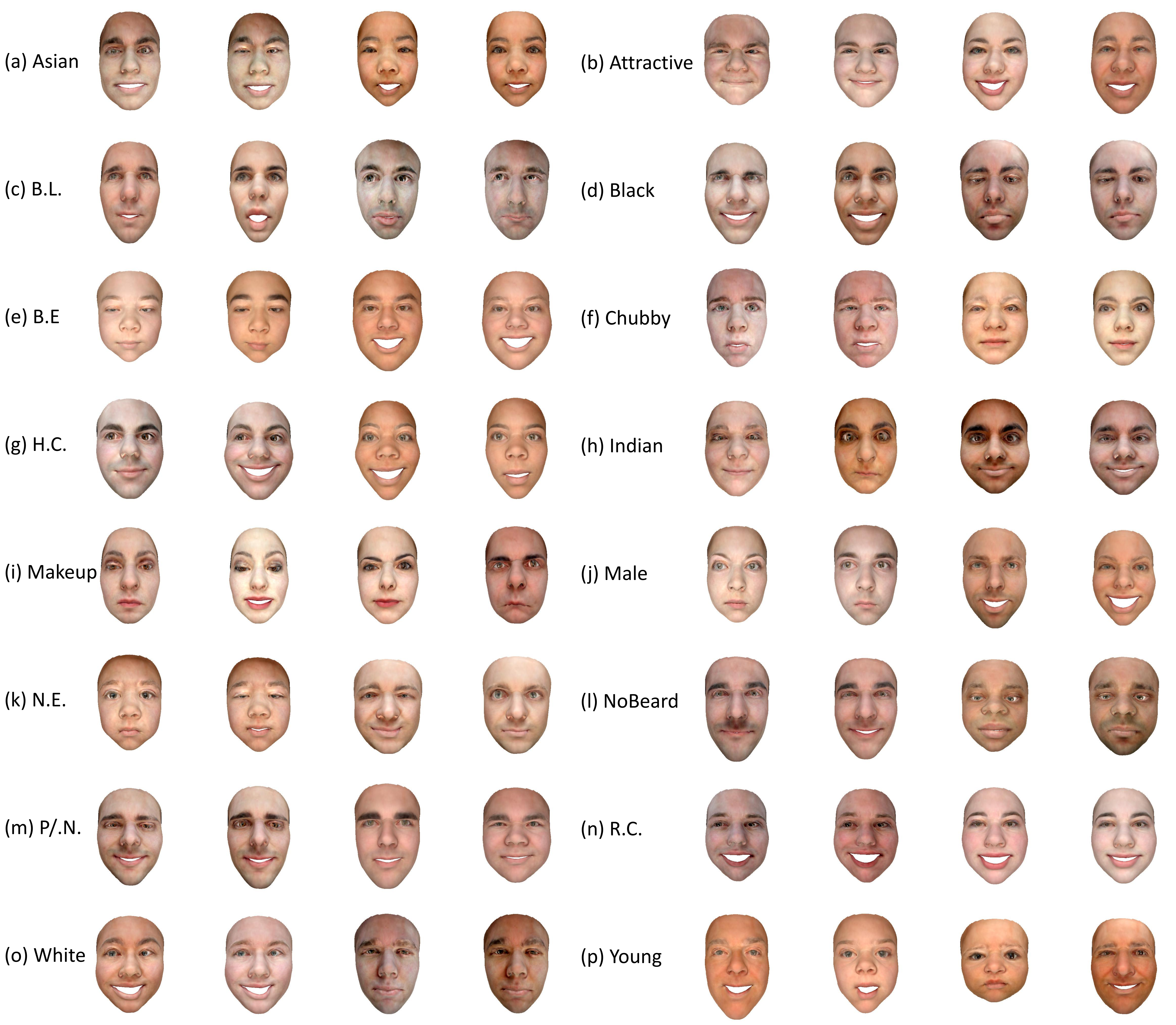}
  \caption{Results of our conditional attribute controller given the FFHQ parameters. For each attribute, the second and fourth columns show the results of transforming the first and the second columns to gain and lose the attribute respectively.}
\label{fig:FFHQ_figure}
\end{figure*}

\vspace{1mm}  \noindent \textbf{Comparisons with the baseline.}
We qualitatively compare our conditional attribute controller and the baseline by using the samples used in Table~\ref{table:l2}.
Both models are learned from the synthetic dataset.
As seen in Fig.~\ref{fig:l2-comparisons}, the outputs of our controller shows more proper semantic changes for a given parameter. The outputs of baseline, however, remain undesirable features, e.g., beard in (c), or contain unnatural characteristics after being semantically manipulated.
For example, the baseline output in (a) shows partial complexion changes and the output in (b) maintains the baby's features of a short face height after being transformed to the `Old' attribute.
We believe those undesirable semantic changes reveal the limitations of the baseline global method. On the other hand, the outputs of ours show more understandable semantic transformations considering a given parameter and a target attribute. We believe this enhancement comes from the conditional training scheme enabled by our novel synthetic dataset.
\begin{figure}
  \includegraphics[width=\linewidth]{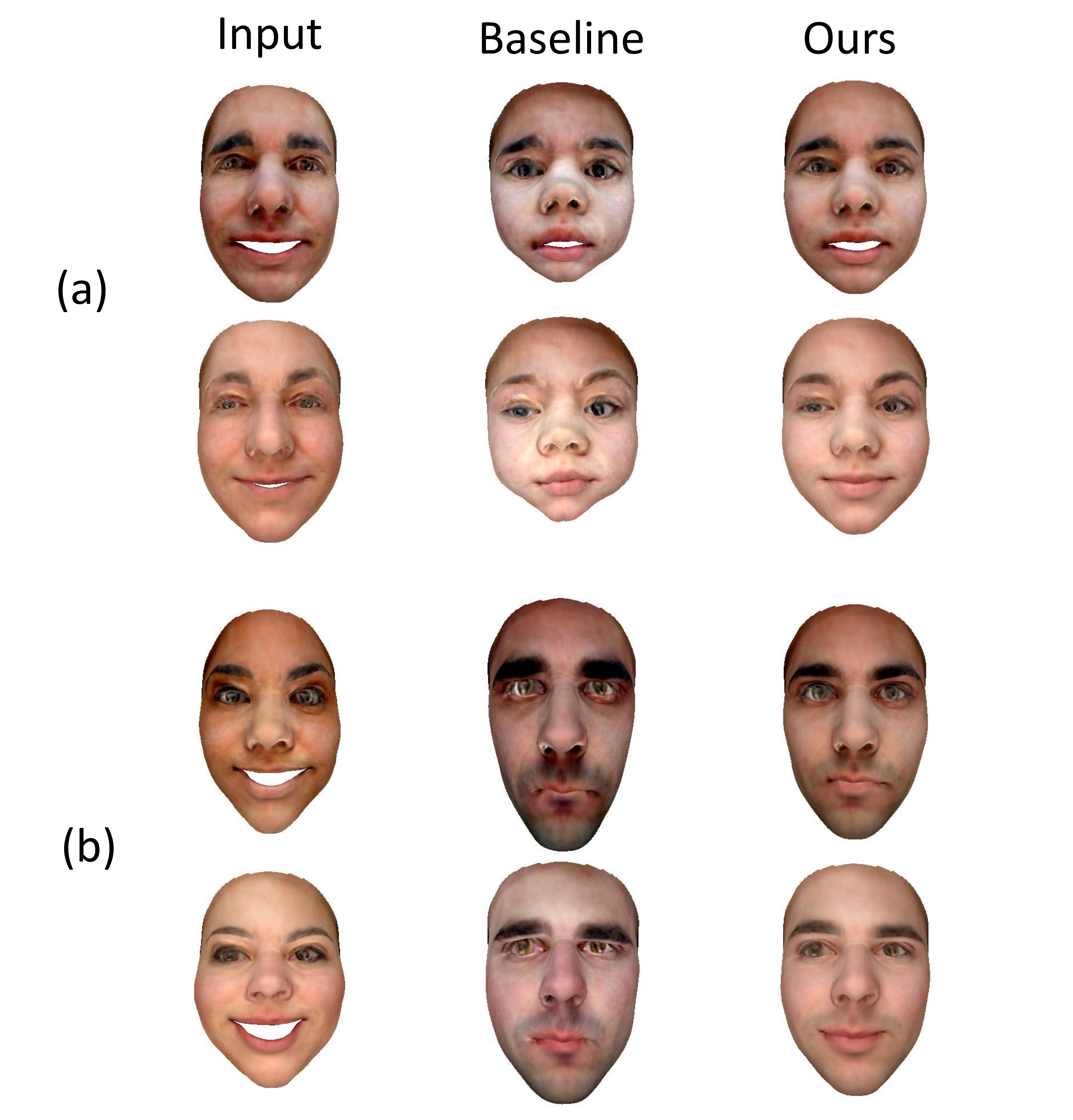}
  \caption{Examples of the same attribute transformations given different input parameters.
  Macro rows respectively indicate (a) Old$\rightarrow{\text{Young}}$ and (b) Female$\rightarrow{\text{Male}}$.}
\label{fig:Comparisons-with-baseline}
\end{figure}
\vspace{1mm}  \noindent \textbf{Benefits of the conditional attribute controller.}
We further compare results from our conditional controller with the ones from the baseline in Fig.~\ref{fig:Comparisons-with-baseline}. The qualitative performance of our method in FFHQ dataset is also demonstrated in Fig.~\ref{fig:FFHQ_figure}.
As seen in Fig.~\ref{fig:Comparisons-with-baseline}, the transformed parameters from our method represent more suitable semantic changes for a given input parameter.
Specifically, the baseline method shows the same transformations for each macro row---i.e., regardless of the input, the transformed 3D faces through baseline consistently represent the shortened face in (a) and the stretched face in (b). On the other hand, the results from our conditional attribute controller show different transformations, which are more proper for the given input parameter.
This experiments justify the necessity of our conditional scheme in 3D facial attribute manipulation. 

\begin{figure}[H]
\vspace*{-0.5cm}
  \includegraphics[width=\linewidth]{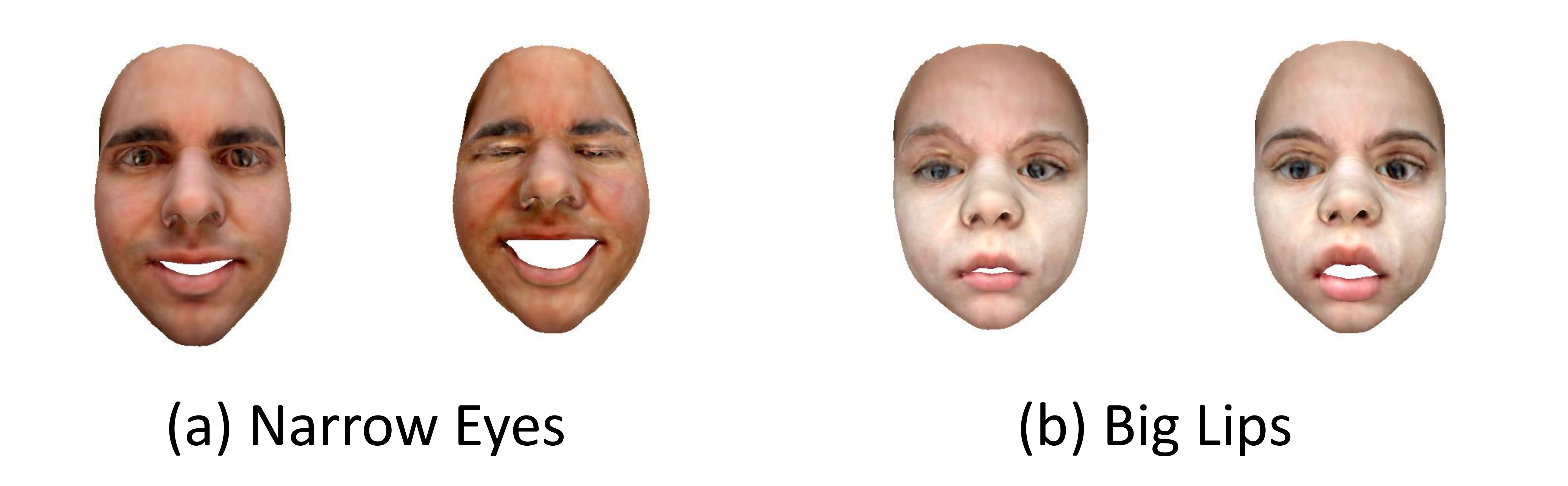}
  \caption{Limitations stem from correlation between attributes.}
\label{fig:limitation}
\vspace*{-0.5cm}
\end{figure}

\vspace{1mm}  \noindent \textbf{Limitations of our method.}
One of the most important components in our pipeline is the GAN latent space.
While exploring the space, we noticed that the semantics in the space are not fully disentangled, i.e., some attributes are highly correlated.
For example, as represented in Fig.~\ref{fig:limitation}, `Narrow Eyes' is closely related to the `Smile' attribute and `Big Lips' is positively correlated with `Opened Mouth'.
However, we expect the disentanglement of the GAN latent space to be improved in the future, and any improvements could easily be incorporated into our approach.

\section{Conclusion}
In this paper, we present a novel 3D dataset creation pipeline and a conditional attribute controller.
The superior performance of a combination of our proposed methods is rigorously demonstrated in our experiments.
The experiments throughout our work verify that the precision and the diversity of the semantically transformed 3DMM parameters are enhanced.
Our work could be readily extended to non-linear 3DMM models provided that the 3D reconstruction part in our pipeline is changed to the non-linear 3DMM model.
We believe our conditional attribute controller could further outperform the global additive baseline when using non-linear 3DMM because translating in the same direction would perform worse for a non-linear parametric space.
We hope our research will broaden the applicability and interest of 3D facial attribute manipulation.





\begin{figure*}[t]
  \includegraphics[width=\linewidth]{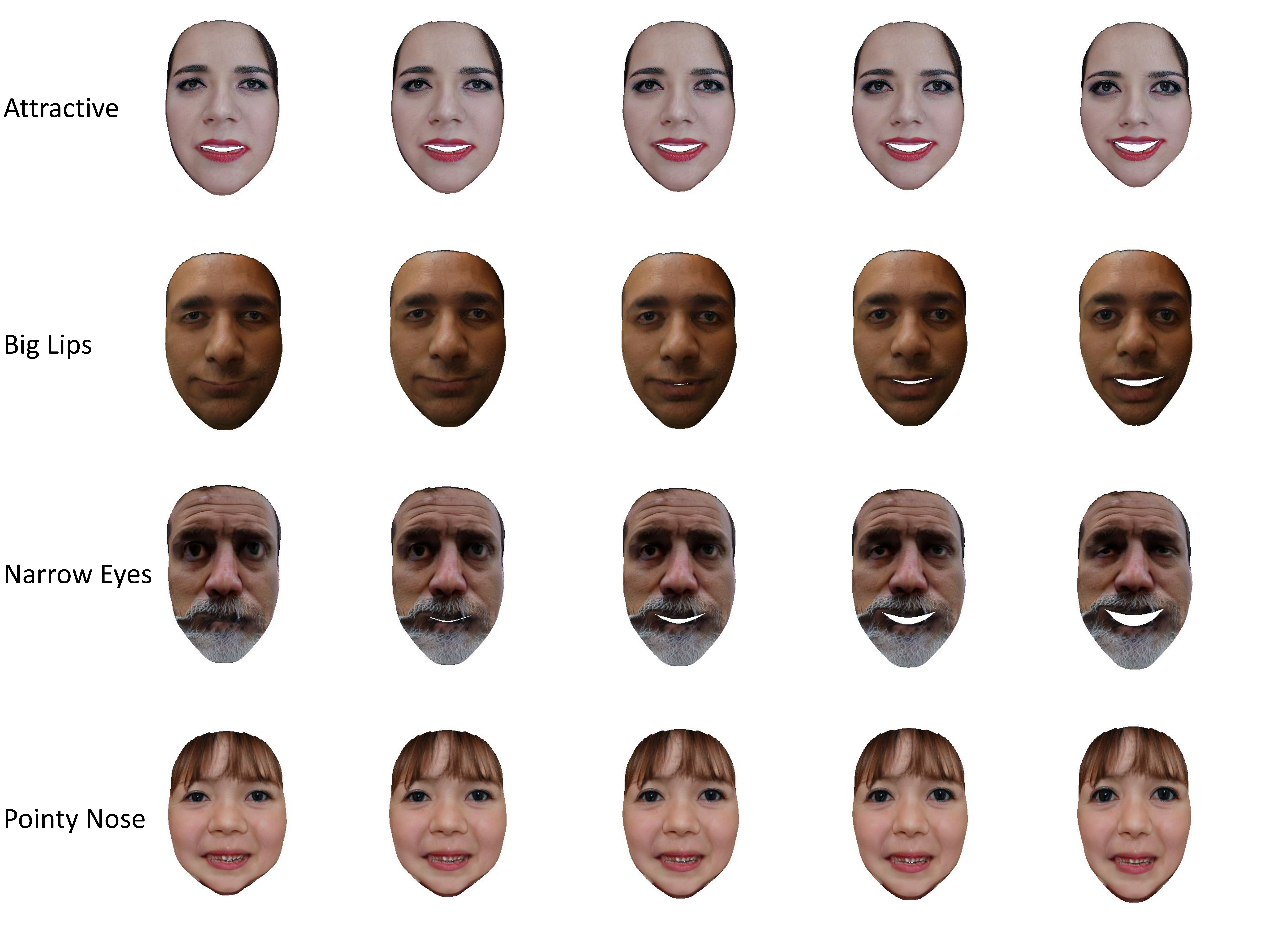}
  \vspace{-1cm}
  \caption{A combination of our conditional attribute controller with a high-quality texture model.} 
\label{fig:application}
\end{figure*}

\section{Appendix}
\vspace{-1mm} 
\noindent \textbf{Applications.}
In order to verify the usefulness of our proposed idea, we report an example case of possible applications. To elaborate, we leverage a high-quality texture model~\cite{styleuv} for the texture representation and use the shape and expression models of 3DMM. By controlling the 3DMM parameters for the shape and the expression through the conditional attribute controller, we semantically manipulate the attributes of the high-quality 3D face. Fig.~\ref{fig:application} verifies that our method can be easily combined with the high-quality texture model, which can be useful in practice.

Furthermore, our method can be readily extended to non-linear models once the linear 3DMM models in our novel pipeline is replaced with the non-linear models. We think this research direction would be an interesting extension of our proposed method. Another possible application is 3DMM-based image manipulation~\cite{tewari2020stylerig,deng2020disentangled,ghosh2020gif}. With our controller, the utility of these methods would significantly increase. This is because the number of attributes the current methods can deal with is limited to the pose, light, and expressions. Once our method is combined with those methods, we believe it is possible to make a lot of attributes controlled, which would be beneficial for a potential user.

\begin{figure*}
  \includegraphics[width=\linewidth]{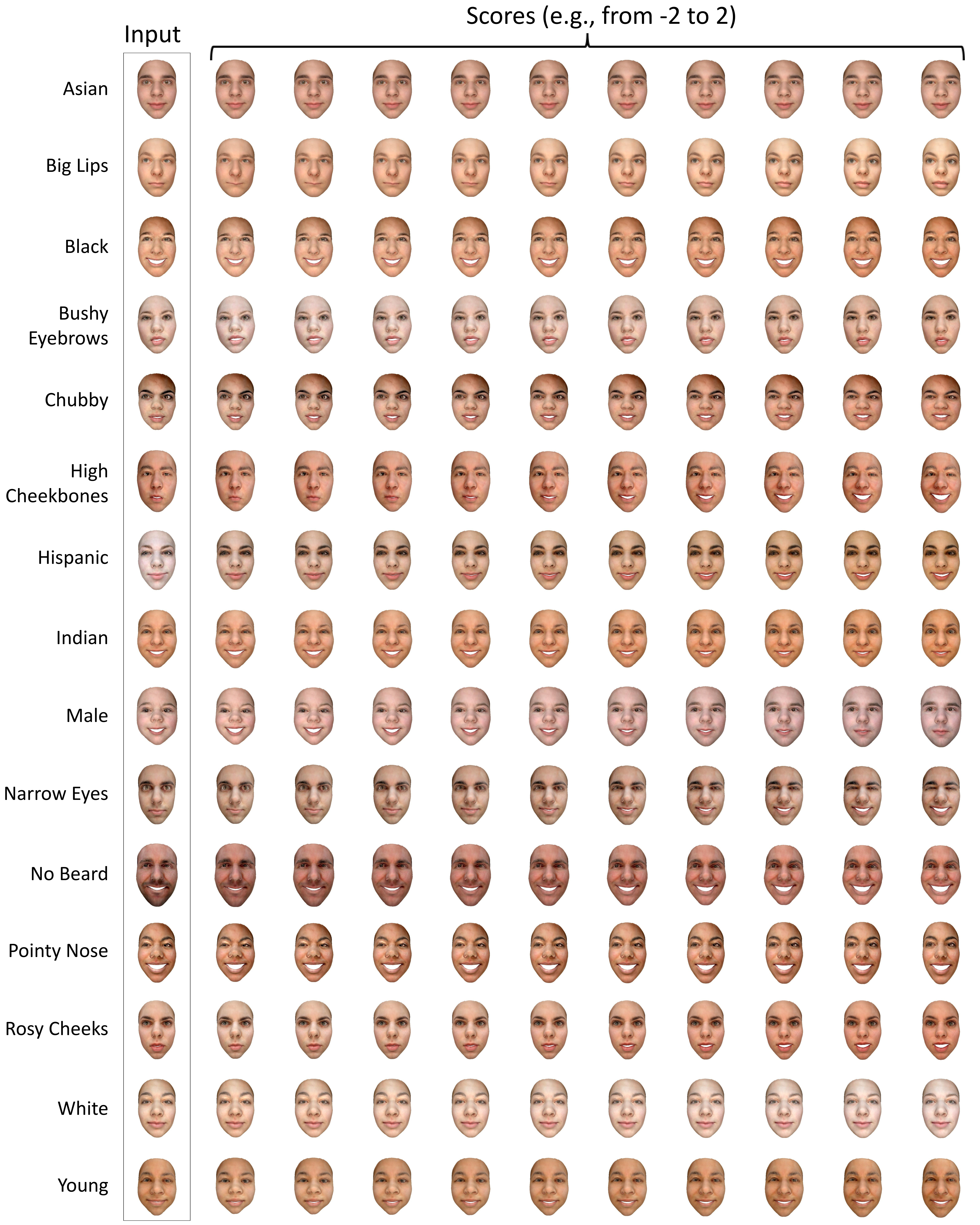}
  \vspace{-0.5cm}
  \caption{Visualization of semantic changes of 3D face given different scores.} 
\label{fig:vis-effects-different-scores}
\vspace*{-0.3cm}
\end{figure*}

\vspace{1mm} 
\noindent \textbf{Visualizations of effects of score.}
Fig.~\ref{fig:vis-effects-different-scores} shows varied semantic transformations of 3D faces according to changes of the score. The first column shows an input 3DMM parameter ${p_{src}}$ and the other columns represent the transformed 3D parameters for the given parameter and the score, i.e., ${p_{src}+f(p_{src},s)}$, where ${s\in\{-2.0,-1.5,-1.0,...,2.0\}}$.

The results verify that our conditional attribute controller, learned from our novel synthetic dataset can perform a proper semantic transformation for the given score. For example, the `Big Lips' attribute in the second row in the figure shows the thicker lips as bigger the score.

We can also observe the limitations of our synthetic data, as noted in subsection~\ref{subsection:qualitative evaluations}. The `Narrow Eyes' attribute in the sixth row from bottom makes not only the eyes of the 3D face closed but also its mouth smiled. Once the disentangling capability of the GANs' latent space is improved, we believe our pipeline can be enhanced as well.

\clearpage
{\small
\bibliographystyle{ieee_fullname}
\bibliography{egbib}

\begin{thebibliography}{10}\itemsep=-1pt

\bibitem{allen2003space}
Brett Allen, Brian Curless, and Zoran Popovi{\'c}.
\newblock The space of human body shapes: reconstruction and parameterization
  from range scans.
\newblock {\em ACM transactions on graphics (TOG)}, 22(3):587--594, 2003.

\bibitem{TheSpaceofHumanBodyShapes}
Brett Allen, Brian Curless, and Zoran Popovi\'{c}.
\newblock The space of human body shapes: Reconstruction and parameterization
  from range scans.
\newblock {\em ACM Trans. Graph.}, 22(3):587–594, July 2003.

\bibitem{WGAN}
Martin Arjovsky, Soumith Chintala, and L\'{e}on Bottou.
\newblock Wasserstein generative adversarial networks.
\newblock In {\em Proceedings of the 34th International Conference on Machine
  Learning - Volume 70}, ICML'17, page 214–223. JMLR.org, 2017.

\bibitem{blanz2003reanimating}
Volker Blanz, Curzio Basso, Tomaso Poggio, and Thomas Vetter.
\newblock Reanimating faces in images and video.
\newblock In {\em Computer graphics forum}, volume~22, pages 641--650. Wiley
  Online Library, 2003.

\bibitem{3DMM}
Volker Blanz and Thomas Vetter.
\newblock A morphable model for the synthesis of 3d faces.
\newblock In {\em Proceedings of the 26th Annual Conference on Computer
  Graphics and Interactive Techniques}, SIGGRAPH '99, page 187–194, USA,
  1999. ACM Press/Addison-Wesley Publishing Co.

\bibitem{Face-recognition-based}
V. {Blanz} and T. {Vetter}.
\newblock Face recognition based on fitting a 3d morphable model.
\newblock {\em IEEE Transactions on Pattern Analysis and Machine Intelligence},
  25(9):1063--1074, 2003.

\bibitem{Blanz_fitting}
Volker Blanz and Thomas Vetter.
\newblock Face recognition based on fitting a 3d morphable model.
\newblock {\em IEEE Trans. Pattern Anal. Mach. Intell.}, 25(9):1063–1074,
  Sept. 2003.

\bibitem{LSFM}
J. {Booth}, A. {Roussos}, S. {Zafeiriou}, A. {Ponniah}, and D. {Dunaway}.
\newblock A 3d morphable model learnt from 10,000 faces.
\newblock In {\em 2016 IEEE Conference on Computer Vision and Pattern
  Recognition (CVPR)}, pages 5543--5552, 2016.

\bibitem{brock2018large}
Andrew Brock, Jeff Donahue, and Karen Simonyan.
\newblock Large scale {GAN} training for high fidelity natural image synthesis.
\newblock In {\em International Conference on Learning Representations}, 2019.

\bibitem{FaceWarehouse}
Chen Cao, Yanlin Weng, Shun Zhou, Yiying Tong, and Kun Zhou.
\newblock Facewarehouse: A 3d facial expression database for visual computing.
\newblock {\em IEEE Transactions on Visualization and Computer Graphics},
  20(3):413–425, Mar. 2014.

\bibitem{chai2003vision}
Jin-xiang Chai, Jing Xiao, and Jessica Hodgins.
\newblock Vision-based control of 3 d facial animation.
\newblock In {\em Symposium on Computer Animation}, volume~1, 2003.

\bibitem{GDWCT2019}
Wonwoong Cho, Sungha Choi, David~Keetae Park, Inkyu Shin, and Jaegul Choo.
\newblock Image-to-image translation via group-wise deep whitening-and-coloring
  transformation.
\newblock In {\em The IEEE Conference on Computer Vision and Pattern
  Recognition (CVPR)}, 2019.

\bibitem{choi2018stargan}
Yunjey Choi, Minje Choi, Munyoung Kim, Jung-Woo Ha, Sunghun Kim, and Jaegul
  Choo.
\newblock Stargan: Unified generative adversarial networks for multi-domain
  image-to-image translation.
\newblock In {\em Proceedings of the IEEE Conference on Computer Vision and
  Pattern Recognition}, 2018.

\bibitem{deng2020disentangled}
Yu Deng, Jiaolong Yang, Dong Chen, Fang Wen, and Xin Tong.
\newblock Disentangled and controllable face image generation via 3d
  imitative-contrastive learning.
\newblock In {\em Proceedings of the IEEE/CVF Conference on Computer Vision and
  Pattern Recognition}, pages 5154--5163, 2020.

\bibitem{deng2019accurate}
Yu Deng, Jiaolong Yang, Sicheng Xu, Dong Chen, Yunde Jia, and Xin Tong.
\newblock Accurate 3d face reconstruction with weakly-supervised learning: From
  single image to image set.
\newblock In {\em IEEE Computer Vision and Pattern Recognition Workshops},
  2019.

\bibitem{History}
Bernhard Egger, William A.~P. Smith, Ayush Tewari, Stefanie Wuhrer, Michael
  Zollhoefer, Thabo Beeler, Florian Bernard, Timo Bolkart, Adam Kortylewski,
  Sami Romdhani, Christian Theobalt, Volker Blanz, and Thomas Vetter.
\newblock 3d morphable face models—past, present, and future.
\newblock {\em ACM Trans. Graph.}, 39(5), June 2020.

\bibitem{BFM17}
T. {Gerig}, A. {Morel-Forster}, C. {Blumer}, B. {Egger}, M. {Luthi}, S.
  {Schoenborn}, and T. {Vetter}.
\newblock Morphable face models - an open framework.
\newblock In {\em 2018 13th IEEE International Conference on Automatic Face
  Gesture Recognition (FG 2018)}, pages 75--82, 2018.

\bibitem{ghafourzadeh2019part}
Donya Ghafourzadeh, Cyrus Rahgoshay, Sahel Fallahdoust, Adeline Aubame, Andre
  Beauchamp, Tiberiu Popa, and Eric Paquette.
\newblock Part-based 3d face morphable model with anthropometric local control.
\newblock 2019.

\bibitem{Ghafourzadeh2019PartBased3F}
Donya Ghafourzadeh, Cyrus Rahgoshay, Sahel Fallahdoust, A. Beauchamp, Adeline
  Aubame, and E. Paquette.
\newblock Part-based 3d face morphable model with anthropometric local control.
\newblock 2019.

\bibitem{ghosh2020gif}
Partha Ghosh, Pravir~Singh Gupta, Roy Uziel, Anurag Ranjan, Michael Black, and
  Timo Bolkart.
\newblock Gif: Generative interpretable faces.
\newblock {\em arXiv preprint arXiv:2009.00149}, 2020.

\bibitem{GAN}
Ian~J. Goodfellow, Jean Pouget-Abadie, Mehdi Mirza, Bing Xu, David
  Warde-Farley, Sherjil Ozair, Aaron Courville, and Yoshua Bengio.
\newblock Generative adversarial nets.
\newblock In {\em Proceedings of the 27th International Conference on Neural
  Information Processing Systems - Volume 2}, NIPS'14, page 2672–2680,
  Cambridge, MA, USA, 2014. MIT Press.

\bibitem{wgan-gp}
Ishaan Gulrajani, Faruk Ahmed, Martin Arjovsky, Vincent Dumoulin, and Aaron~C
  Courville.
\newblock Improved training of wasserstein gans.
\newblock In I. Guyon, U.~V. Luxburg, S. Bengio, H. Wallach, R. Fergus, S.
  Vishwanathan, and R. Garnett, editors, {\em Advances in Neural Information
  Processing Systems}, volume~30, pages 5767--5777. Curran Associates, Inc.,
  2017.

\bibitem{huang2018munit}
Xun Huang, Ming-Yu Liu, Serge Belongie, and Jan Kautz.
\newblock Multimodal unsupervised image-to-image translation.
\newblock In {\em ECCV}, 2018.

\bibitem{ganspace}
Erik Härkönen, Aaron Hertzmann, Jaakko Lehtinen, and Sylvain Paris.
\newblock Ganspace: Discovering interpretable gan controls.
\newblock {\em CoRR}, abs/2004.02546, 2020.

\bibitem{bfm09}
IEEE.
\newblock {\em A 3D Face Model for Pose and Illumination Invariant Face
  Recognition}, Genova, Italy, 2009.

\bibitem{Large-Pose-Face-Alignment}
A. {Jourabloo} and X. {Liu}.
\newblock Large-pose face alignment via cnn-based dense 3d model fitting.
\newblock In {\em 2016 IEEE Conference on Computer Vision and Pattern
  Recognition (CVPR)}, pages 4188--4196, 2016.

\bibitem{karras2018progressive}
Tero Karras, Timo Aila, Samuli Laine, and Jaakko Lehtinen.
\newblock Progressive growing of gans for improved quality, stability, and
  variation.
\newblock In {\em International Conference on Learning Representations}, 2018.

\bibitem{Karras_2019_CVPR}
Tero Karras, Samuli Laine, and Timo Aila.
\newblock A style-based generator architecture for generative adversarial
  networks.
\newblock In {\em Proceedings of the IEEE/CVF Conference on Computer Vision and
  Pattern Recognition (CVPR)}, June 2019.

\bibitem{Kazemi_2014_CVPR}
Vahid Kazemi and Josephine Sullivan.
\newblock One millisecond face alignment with an ensemble of regression trees.
\newblock In {\em Proceedings of the IEEE Conference on Computer Vision and
  Pattern Recognition (CVPR)}, June 2014.

\bibitem{kim2018deep}
Hyeongwoo Kim, Pablo Garrido, Ayush Tewari, Weipeng Xu, Justus Thies, Matthias
  Niessner, Patrick P{\'e}rez, Christian Richardt, Michael Zollh{\"o}fer, and
  Christian Theobalt.
\newblock Deep video portraits.
\newblock {\em ACM Transactions on Graphics (TOG)}, 37(4):1--14, 2018.

\bibitem{gaussian_mixture_3dmm}
Paul Koppen, Zhen-Hua Feng, Josef Kittler, Muhammad Awais, William Christmas,
  Xiao-Jun Wu, and He-Feng Yin.
\newblock Gaussian mixture 3d morphable face model.
\newblock {\em Pattern Recogn.}, 74(C):617–628, Feb. 2018.

\bibitem{styleuv}
Myunggi Lee, Wonwoong Cho, Moonheum Kim, David Inouye, and Nojun Kwak.
\newblock Styleuv: Diverse and high-fidelity uv map generative model.
\newblock In {\em arXiv preprint}, 2020.

\bibitem{lewis2010direct}
John~P Lewis and Ken-ichi Anjyo.
\newblock Direct manipulation blendshapes.
\newblock {\em IEEE Computer Graphics and Applications}, 30(4):42--50, 2010.

\bibitem{1613022}
{Lijun Yin}, {Xiaozhou Wei}, {Yi Sun}, {Jun Wang}, and M.~J. {Rosato}.
\newblock A 3d facial expression database for facial behavior research.
\newblock In {\em 7th International Conference on Automatic Face and Gesture
  Recognition (FGR06)}, pages 211--216, 2006.

\bibitem{liu2019softras}
Shichen Liu, Tianye Li, Weikai Chen, and Hao Li.
\newblock Soft rasterizer: A differentiable renderer for image-based 3d
  reasoning.
\newblock {\em The IEEE International Conference on Computer Vision (ICCV)},
  Oct 2019.

\bibitem{liu2015faceattributes}
Ziwei Liu, Ping Luo, Xiaogang Wang, and Xiaoou Tang.
\newblock Deep learning face attributes in the wild.
\newblock In {\em Proceedings of International Conference on Computer Vision
  (ICCV)}, December 2015.

\bibitem{NEURIPS2018_e46de7e1}
Mario Lucic, Karol Kurach, Marcin Michalski, Sylvain Gelly, and Olivier
  Bousquet.
\newblock Are gans created equal? a large-scale study.
\newblock In S. Bengio, H. Wallach, H. Larochelle, K. Grauman, N. Cesa-Bianchi,
  and R. Garnett, editors, {\em Advances in Neural Information Processing
  Systems}, volume~31, pages 700--709. Curran Associates, Inc., 2018.

\bibitem{Mescheder2018ICML}
Lars Mescheder, Sebastian Nowozin, and Andreas Geiger.
\newblock Which training methods for gans do actually converge?
\newblock In {\em International Conference on Machine Learning (ICML)}, 2018.

\bibitem{miyato2018spectral}
Takeru Miyato, Toshiki Kataoka, Masanori Koyama, and Yuichi Yoshida.
\newblock Spectral normalization for generative adversarial networks.
\newblock In {\em International Conference on Learning Representations}, 2018.

\bibitem{neumann2013sparse}
Thomas Neumann, Kiran Varanasi, Stephan Wenger, Markus Wacker, Marcus Magnor,
  and Christian Theobalt.
\newblock Sparse localized deformation components.
\newblock {\em ACM Transactions on Graphics (TOG)}, 32(6):1--10, 2013.

\bibitem{niswar2011virtual}
Arthur Niswar, Ishtiaq~Rasool Khan, and Farzam Farbiz.
\newblock Virtual try-on of eyeglasses using 3d model of the head.
\newblock In {\em Proceedings of the 10th International Conference on Virtual
  Reality Continuum and Its Applications in Industry}, pages 435--438, 2011.

\bibitem{phong}
Bui~Tuong Phong.
\newblock Illumination for computer generated pictures.
\newblock {\em Commun. ACM}, 18(6):311–317, June 1975.

\bibitem{Romdhani05estimating3d}
Sami Romdhani and Thomas Vetter.
\newblock Estimating 3d shape and texture using pixel intensity, edges,
  specular highlights, texture constraints and a prior.
\newblock In {\em Edges, Specular Highlights, Texture Constraints and a Prior,
  Proceedings of Computer Vision and Pattern Recognition}, pages 986--993,
  2005.

\bibitem{Estimating3DShape}
Sami Romdhani and Thomas Vetter.
\newblock Estimating 3d shape and texture using pixel intensity, edges,
  specular highlights, texture constraints and a prior.
\newblock In {\em Proceedings of the 2005 IEEE Computer Society Conference on
  Computer Vision and Pattern Recognition (CVPR'05) - Volume 2 - Volume 02},
  CVPR '05, page 986–993, USA, 2005. IEEE Computer Society.

\bibitem{scherbaum2011computer}
Kristina Scherbaum, Tobias Ritschel, Matthias Hullin, Thorsten Thorm{\"a}hlen,
  Volker Blanz, and Hans-Peter Seidel.
\newblock Computer-suggested facial makeup.
\newblock In {\em Computer Graphics Forum}, volume~30, pages 485--492. Wiley
  Online Library, 2011.

\bibitem{Shen_2020_CVPR}
Yujun Shen, Jinjin Gu, Xiaoou Tang, and Bolei Zhou.
\newblock Interpreting the latent space of gans for semantic face editing.
\newblock In {\em Proceedings of the IEEE/CVF Conference on Computer Vision and
  Pattern Recognition (CVPR)}, 2020.

\bibitem{shen2020closedform}
Yujun Shen and Bolei Zhou.
\newblock Closed-form factorization of latent semantics in gans.
\newblock {\em arXiv preprint arXiv:2007.06600}, 2020.

\bibitem{Bodytalk}
Stephan Streuber, M.~Alejandra Quiros-Ramirez, Matthew~Q. Hill, Carina~A. Hahn,
  Silvia Zuffi, Alice O'Toole, and Michael~J. Black.
\newblock Body talk: Crowdshaping realistic 3d avatars with words.
\newblock {\em ACM Trans. Graph.}, 35(4), July 2016.

\bibitem{deepface}
Y. {Taigman}, M. {Yang}, M. {Ranzato}, and L. {Wolf}.
\newblock Deepface: Closing the gap to human-level performance in face
  verification.
\newblock In {\em 2014 IEEE Conference on Computer Vision and Pattern
  Recognition}, pages 1701--1708, 2014.

\bibitem{tena2011interactive}
J~Rafael Tena, Fernando De~la Torre, and Iain Matthews.
\newblock Interactive region-based linear 3d face models.
\newblock In {\em ACM SIGGRAPH 2011 papers}, pages 1--10, 2011.

\bibitem{tewari2020stylerig}
Ayush Tewari, Mohamed Elgharib, Gaurav Bharaj, Florian Bernard, Hans-Peter
  Seidel, Patrick P{\'e}rez, Michael Zollhofer, and Christian Theobalt.
\newblock Stylerig: Rigging stylegan for 3d control over portrait images.
\newblock In {\em Proceedings of the IEEE/CVF Conference on Computer Vision and
  Pattern Recognition}, pages 6142--6151, 2020.

\bibitem{tewari2017mofa}
Ayush Tewari, Michael Zollhofer, Hyeongwoo Kim, Pablo Garrido, Florian Bernard,
  Patrick Perez, and Christian Theobalt.
\newblock Mofa: Model-based deep convolutional face autoencoder for
  unsupervised monocular reconstruction.
\newblock In {\em Proceedings of the IEEE International Conference on Computer
  Vision Workshops}, pages 1274--1283, 2017.

\bibitem{face-transfer}
Daniel Vlasic, Matthew Brand, Hanspeter Pfister, and Jovan Popovic.
\newblock Face transfer with multilinear models.
\newblock In {\em ACM SIGGRAPH 2006 Courses}, SIGGRAPH '06, page 24–es, New
  York, NY, USA, 2006. Association for Computing Machinery.

\bibitem{voynov2020unsupervised}
Andrey Voynov and Artem Babenko.
\newblock Unsupervised discovery of interpretable directions in the gan latent
  space.
\newblock In {\em ICML}, 2020.

\bibitem{icml2020_2025}
Andrey Voynov and Artem Babenko.
\newblock Unsupervised discovery of interpretable directions in the gan latent
  space.
\newblock In {\em Proceedings of Machine Learning and Systems 2020}, pages
  3515--3525, 2020.

\bibitem{zhifei2017cvpr}
Song Zhang, Zhifei, Yang, and Hairong Qi.
\newblock Age progression/regression by conditional adversarial autoencoder.
\newblock In {\em IEEE Conference on Computer Vision and Pattern Recognition
  (CVPR)}. IEEE, 2017.

\bibitem{CycleGAN2017}
Jun-Yan Zhu, Taesung Park, Phillip Isola, and Alexei~A Efros.
\newblock Unpaired image-to-image translation using cycle-consistent
  adversarial networks.
\newblock In {\em Computer Vision (ICCV), 2017 IEEE International Conference
  on}, 2017.

\bibitem{zhu2016face}
Xiangyu Zhu, Zhen Lei, Xiaoming Liu, Hailin Shi, and Stan~Z Li.
\newblock Face alignment across large poses: A 3d solution.
\newblock In {\em Proceedings of the IEEE conference on computer vision and
  pattern recognition}, pages 146--155, 2016.

\end{thebibliography}
}

\end{document}